\documentclass[10pt,letterpaper]{article}

\usepackage{cogsci}
\usepackage{pslatex}
\usepackage{apacite}
\usepackage{graphicx}
\usepackage[small]{caption}
\usepackage[compact]{titlesec}
\usepackage{url}
\usepackage{amssymb}
\usepackage{amsmath}
\usepackage{todonotes}

\titlespacing{\section}{0pt}{*0.5}{*0.5}
\titlespacing{\subsection}{0pt}{*0.5}{*0.5}
\titlespacing{\subsubsection}{0pt}{*0.1}{*1}

\title{Physical problem solving: \\Joint planning with symbolic, geometric, and dynamic constraints\vspace{-0.2cm}}

\author{{\large {\bf Ilker Yildirim*$^1$} (ilkery@mit.edu)}, {\large {\bf Tobias Gerstenberg*$^1$} (tger@mit.edu)}, {\large {\bf Basil Saeed$^1$} (bsaeed@mit.edu)},\\ {\large {\bf Marc Toussaint$^{2}$} (marc.toussaint@informatik.uni-stuttgart.de)}, {\large {\bf Joshua B. Tenenbaum$^1$} (jbt@mit.edu)}\\
$^1$ Brain and Cognitive Sciences, Massachusetts Institute of Technology, Cambridge, MA \\ $^2$ Machine Learning and Robotics Lab, University of Stuttgart, Germany
\vspace{-0.5cm}
}

\begin{document}
\maketitle
\begin{abstract}
\vspace{-0.1cm}

In this paper, we present a new task that investigates how people interact with and make judgments about towers of blocks. In Experiment~1, participants in the lab solved a series of problems in which they had to re-configure three blocks from an initial to a final configuration. We recorded whether they used one hand or two hands to do so. In Experiment~2, we asked participants online to judge whether they think the person in the lab used one or two hands. The results revealed a close correspondence between participants' actions in the lab, and the mental simulations of participants online. To explain participants' actions and mental simulations, we develop a model that plans over a symbolic representation of the situation, executes the plan using a geometric solver, and checks the plan's feasibility by taking into account the physical constraints of the scene. Our model explains participants' actions and judgments to a high degree of quantitative accuracy. 

\end{abstract}

\vspace{-0.3cm}
\section{Introduction}


Physical problem solving -- converting knowledge into behavior to achieve a goal that involves physical object manipulation -- is a core component of human intelligence and ubiquitous in everyday cognition.  From young children playing with stacking cups to an adult moving furniture to redesign a room or to load a truck, our intuitive understanding of how to manipulate the physical world in order to meet our goals is remarkable. For instance, when rearranging the furniture in a room, one needs to form and execute a plan which takes into account both spatial and physical constraints, such as how big are the objects, and which objects might be stacked on top of others. 

Two independently developed lines of research provide insights and starting points into exploring these computations: reasoning based on mental models, and motor control based on forward models. Firstly, the theoretical and behavioral work on reasoning and problem solving in symbolic domains (e.g., logical reasoning, or visuo-spatial reasoning) emphasizes the importance of common-sense knowledge. For instance, early Artificial Intelligence (AI) systems that were built to reason like humans do, focused on building models that capture aspects of common-sense knowledge about the physical world in the form of knowledge representations and methods to efficiently manipulate them~\cite<e.g.,>{newell1958elements}. Similarly, in cognitive psychology, the idea that problem solving begins with the construction of a mental model of the situation was explored in more detail by mental model theory \cite{johnson2005mental}. While still operating over logical representations, mental model theory makes additional assumptions about what aspects of a situation people naturally represent, and how these representations support reasoning \cite{johnson-laird2015logic-pr}. However, the theoretical and behavioral work on human reasoning and problem solving has tended to focus on symbolic domains~\cite<e.g., logical, spatial, and visuo-spatial reasoning>{newman2003frontal, byrne1989spatial}, and has not yet looked into situations that require reasoning about physical objects, and forming plans about how to interact with them.

Secondly, research on computational motor control and object manipulation emphasizes the knowledge and transformations necessary for skillful manipulation of objects. For instance, work on sensorimotor control and object manipulation extensively studied internal models of the forward dynamics of the arm and the objects, as well as how to choose actions to efficiently achieve one's goals based on internal models~\cite{nagengast2009optimal, franklin2011computational}. However, this line of work has tended to focus on relatively simple actions, instead of settings that involve planning longer sequences of moves.

\begin{figure*}[ht!]
\centering
\includegraphics[width=0.8\linewidth]{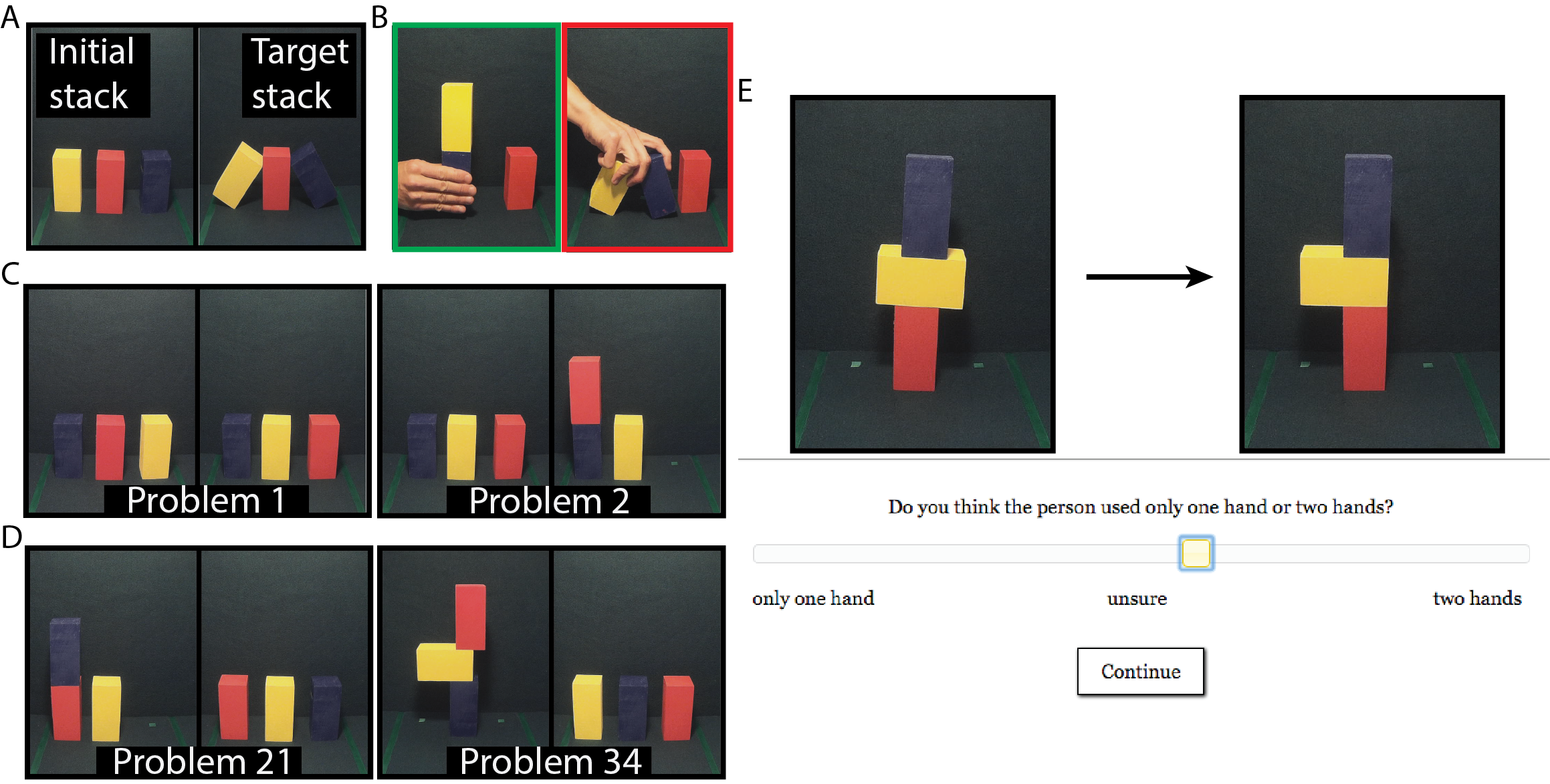}
\vspace{-0.2cm}
\caption{Experimental setup. A: Example for an initial and final configuration of the three blocks. B: Illustration for what moves were legal (green border) or illegal (red border). C and D: Some example problems. E: Screenshot of the experimental interface for participants in Experiment~2.}
\label{fig:experiment}
\vspace{-0.6cm}
\end{figure*}

In this paper, we aim to bring these two different research traditions together. To better understand physical problem solving, we introduce an intuitive, yet complex task in which participants are asked to manipulate a stack of blocks to generate a target configuration. Consider Problem 1 shown in Figure~\ref{fig:experiment}. The task is to manipulate the blocks so that the scene on the left is turned to the scene on the right. While participants have no trouble doing this task, and even young children naturally perform such tasks, modeling people's actions is far from trivial and robotic systems rarely implement this kind of flexible manipulation. The task requires representing the initial state, the final state, and making a plan for how to get from A to B. Finding good action sequences in this task not only requires a symbolic high-level plan (e.g., which sequence of actions to take) and visuo-spatial reasoning, it also requires intuitive physical reasoning about how objects support each other (i.e., their dynamics) and actual motor control required to execute the high-level abstract plan. Such combination of rich behavior is common in everyday cognition, but has rarely been studied in the lab. We used two different versions of the task. In one version, participants in the lab were asked to generate the different configurations. In another version of the task, we had online participants judge whether they think the person in the lab used one or two hands to get from A to B (cf. Figure~\ref{fig:experiment}E). 

We develop a novel computational model of physical problem solving that goes all the way from formulating an abstract symbolic plan to executing the low-level motor commands that are required to realize the plan.
The model is composed of three components: \textit{(1)} a symbolic representation of the scene, \textit{(2)} a geometric solver for motion synthesis, and \textit{(3)} a physics engine for physical reasoning. Planning in the model operates over the symbolic representation of the scene. Each plan is composed of subgoals and finds a sequence of moves that turn the initial into the final configuration (see, e.g., Figure~\ref{fig:model}C, left side). An optimization-based kinematics solver takes the symbolic plan as its input and generates a full motion plan which we implement in a simulated two-armed robot (Figure~\ref{fig:model}C, right side). We use a physics engine to check whether the plan that the kinematic solver came up with is feasible. More specifically, we test at each point when a subgoal is reached, whether the configuration is physically stable. If the plan includes an unstable configuration, it is discarded (Figure~\ref{fig:model}D for a plan that includes an unstable state). The model's task is to get from the initial stack shown in A to the target stack. However, just taking the red block and moving it to the right so that it's correctly positioned relative to the yellow block, causes the blocks to fall over.


For each pair of initial and target stack, the model is able to generate plans using either only one arm, or both arms. We score each plan based on its efficiency which is a function of the number of the moves it takes to get from the initial to the target stack, as well as the effort that the plan takes. We evaluate the contributions of the three different components of our model through lesion studies (i.e. we remove parts of the model and see how well it does, in order to gauge what components are necessary to capture people's behavior). 


The remainder of this paper is organized as follows: first, we describe a novel, physical problem-solving task and show how participants solve the task in the lab and online. Next, we describe our computational model and analyze how well it does in accounting for participants' behavior. We conclude by highlighting the key contributions of the paper, and by suggesting several lines of future research.

\section{Stack re-configuration problems}

\begin{figure*}[t]
\centering
\includegraphics[width=.8\linewidth]{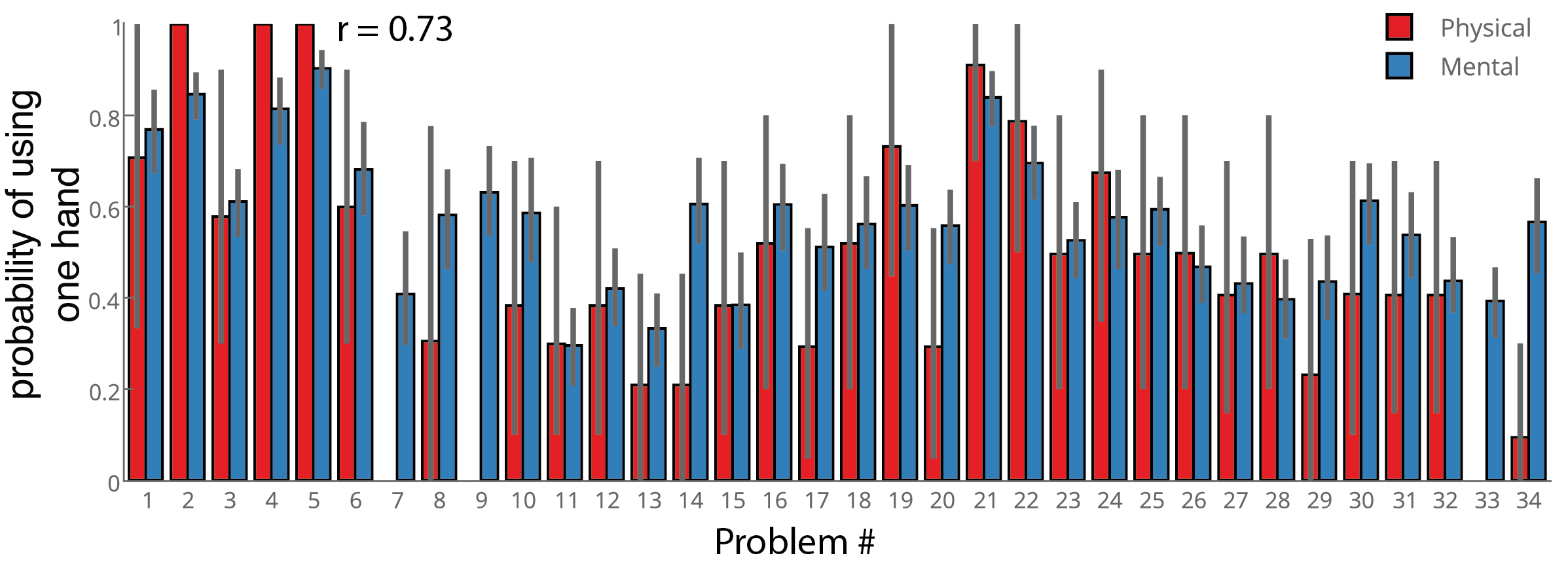}
\vspace{-0.3cm}
\caption{The probability that participants used one hand in the lab (Physical) together with the mean judgments provided by participants online (Mental) for 34 different problems. \emph{Note}: Error bars indicate 95\% bootstrapped confidence intervals.}
\label{fig:expresults}
\vspace{-0.6cm}
\end{figure*}

Most classical paradigms used to study problem solving, such as the Tower of Hanoi and its variants require visuo-spatial reasoning and planning for successful solutions. Here we present a novel problem which requires the problem-solver to also take into account physical constraints, such as considering whether a particular configuration of blocks will be stable. 

The problems involve an initial stack of three physical blocks on a table paired with an image showing the desired target stack of the same three blocks (Figure~\ref{fig:experiment}A). The three wooden blocks had the same size and mass, and were colored in red, yellow, and blue. Given the pair of initial and target stacks, the problem is to re-configure the initial stack such that it will match the target stack in the image. While interacting with the blocks, participants aren't allowed to touch more than one block at a time. Example legal and illegal moves are shown in Fig~\ref{fig:experiment}B. To solve each stack re-configuration problem, participants have to plan and execute a set of moves (using one or both hands) that will generate the target stack from the initial stack. 

\section{Experiment 1: Physical task}
The goal of Experiment~1 was to assess how participants interact with the scene to get from the initial to the final configuration for each problem. In particular, we were interested in seeing whether they used one hand or two hands to get from A to B. 


\subsection{Methods}

\subsubsection{Participants}
10 participants ($M_{\text{age}} = 35, \emph{SD}_{\text{age}} = 16.4, N_{\text{female}} = 6$) were recruited from MIT's subject pool.
The study took about 15 minutes to complete, and all participants were compensated for their participation. 

\subsubsection{Stimuli}
The three physical blocks used in the experiment were of size 10cm-5cm-5cm (height-width-depth) and weighed about 50 grams. We manually arranged these 3 blocks into 38 different configurations and took a picture of each configuration. The configurations were constrained such that all blocks remained within a spatial boundary on a table, and the block or blocks touching the table were centered at one of three designated spots. Figure~\ref{fig:expresults} shows some examples of initial and final configurations.\footnote{For the full set of problems as well as example videos for how the model described below solves the different trials please see: \url{https://github.com/iyildirim/stack-reconfiguration-problems}}


\subsubsection{Procedure}


After providing written consent, participants were introduced to the task, including what moves were legal and which ones were illegal. 
Starting from the initial stack configuration of Problem 1, participants were asked to re-configure the blocks to the target stack of Problem 1, which was presented on a computer screen in front of them. They clicked on the ``Continue'' button on the screen to indicate that they were done and the experiment moved on to the next problem. 

The initial configuration of the next problem, Problem 2 (Figure~\ref{fig:expresults}C), was the target configuration from the previous problem, and so on. 
This sequence of problems continued for a total number of 37 problems.\footnote{Because several participants had trouble to successfully generate the trials 35--37, we will focus on the first 34 trials.} The presentation order was the same for all participants. All participant responses were video-recorded. For each problem, we coded whether participants used one or two hands to solve it.


\subsection{Results}

Figure~\ref{fig:expresults} shows the proportion of participants who used one hand for each trial.  
In some trials, most participants used only one hand (e.g., Problem 21, Figure~\ref{fig:experiment}D), and in others most participants used both hands (e.g., Problem 34, Fig~\ref{fig:experiment}D). Across all trials, participants used one or two hands about equally. 
Participants often solved the problem with one hand if it was possible to do so. 
Some participants only used their non-dominant hand if it was impossible to achieve the target configuration with one hand only.


\subsection{Discussion}

Overall, we found that participants had no trouble doing the task. There was considerable variance in how participants solved the different problems with some participants almost exclusively using one hand (if possible) and others being more likely to use two hands to get to the target configuration.

Experiment~1 serves as a baseline to see how participants actually interact with the physical scene. In Experiment~2, we were interested to see how people mentally simulate the way in which they would interact with the scene to get from the initial to the final stack. If participants are able to mentally do this task, we would expect a close correspondence between the judgments participants make based on their mental simulation, and the actual behavior of participants in the lab.

\section{Experiment 2: Mental task}



The goal of this experiment was to test whether participants can simulate how another person would interact with a physical scene to get from A to B. 


\subsection{Methods}

\subsubsection{Participants}
40 participants ($M_{\text{age}} = 35, \emph{SD}_{\text{age}} = 14, N_{\text{female}} = 22$) were recruited via Amazon's crowdsourcing service Mechanical Turk. The experiment took 8.7 minutes ($SD = 4.4$) to complete and participants were compensated at an hourly rate of $6.0\$$. 

\subsubsection{Stimuli}
The same pairs of initial and target stacks as Experiment 1 were used, with the exception that both stacks were presented on the screen side by side.

\subsubsection{Procedure}

Participants saw two images side by side with the left image showing the initial stack and the right image showing the target stack (example pairs in Figure~\ref{fig:experiment} except panel B).
They were instructed that ``The image on the left shows you the initial configuration of the blocks. The image on the right shows you the configuration after the person interacted with the blocks.'' Their task was to judge whether the person had used one hand or two hands to re-configure the stack. They entered their response by adjusting a slider bar at the bottom of the screen (see Figure~\ref{fig:experiment}). Then they clicked on the ``Continue'' button to proceed to the next problem. The different problems were presented in randomized order. 

\subsection{Results}
Figure~\ref{fig:expresults} shows participants' mean judgments for the different problems. To assess how well participants' mental simulations correspond with the actions that participants took in the experiment, we compared the mean responses in Experiment~2 with the proportion of participants who used one hand in Experiment~1. 



Overall, we found that participants' judgments about how many hands the person used correlated well with participants' actual behavior in the lab, $r=.73$, $p<.05$. Whereas there were many trials for which the correspondence between judgments and actions was very high (e.g. Problems 1--6, or 21--32), there were also situations in which actions and judgments came apart. For example, in Problem~34 almost all participants in the lab used two hands, whereas online participants believed that it was likely that a person would only use one hand to re-configure the scene. 


 

\section{Model}


\begin{figure*}[t]
\centering
\includegraphics[width=.68\linewidth]{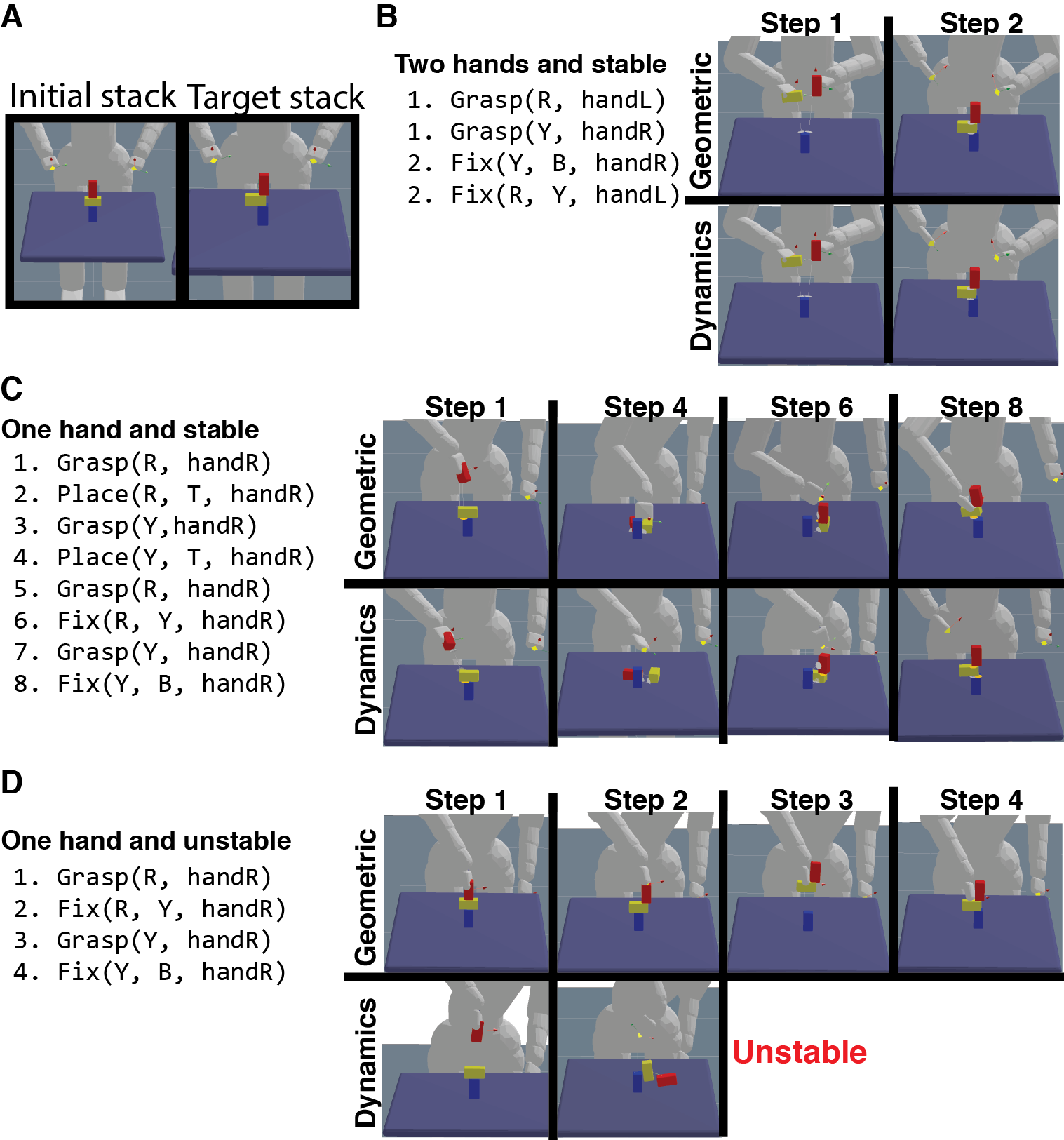}
\vspace{-0.2cm}
\caption{Illustration of how the model works. A: The model successfully went from the inital to the final configuration. B: The symbolic plan for going from Step 1 to Step 2 using two hands. C: A more involved plan that requires 8 moves. D: Example of a scene where a plan fails because it created an unstable configuration (as determined by the physics engine).}
\label{fig:model}
\vspace{-0.6cm}
\end{figure*}

The model consists of three components: \textit{(1)} a set of abstract motion primitives that can be composed to symbolic plans for re-configuring an input stack to a target stack, \textit{(2)} a hierarchical kinematics-based optimization algorithm to find manipulation trajectories conditioned on the symbolic plan, \textit{(3)} and a physics engine to evaluate the stability of the intermediate stages produced by the execution of the manipulation trajectories. The first two components of our model are based on the logic-geometric programming framework~\cite{toussaint2015logic}.

\subsection{Logic-geometric programming framework}

The logic-geometric programming framework presents a solution to problems of combined task and motion planning. Such tasks involve sequential manipulation of a scene based on a geometrically defined goal function. It utilizes symbolic task descriptions as (in-)equality constraints within a hierarchical geometric solver to find full manipulation and object trajectories starting from a coarse-level solution to eventually fine-grained full-paths. Below, we present our representations and an algorithm for symbolic planning as well as a general outline of the geometric solver.

\subsubsection{Symbolic plans}


Symbolic plans are sequences of a set of abstract move types defined using actuators, movable objects and fixed objects in a simulated world. The moves change the state of the actuators and the movable objects. The world is described as a linked list of fixed and movable objects with relative world coordinates: the position and rotation of a child object is defined relative to its parent. 

In order to model our stack re-configuration tasks, we populated the world with three movable objects (red block \verb|R|, green block \verb|G|, and blue block \verb|B|), and a fixed object (table \verb|T|). The world also includes a robotic body with arms and pincer hands (actuators: \verb|handL| and \verb|handR|) overall consisting of 12 degrees of freedom (two at each shoulder, two at each wrist, and two at each hand). 

There are three types of moves: \verb|Grasp(Obj, Act)| specifies a grasp action with an actuator on a movable object. For example, \verb|Grasp(R, handR)| specifies a right hand grasp of the red block. This move changes the position of the object to inside in the actuator while clearing its previous location for moving other objects. The symbolic planning stage doesn't take into account rotation of the objects or the actuators.

\verb|Place(Obj, Supp_Obj, Act)| specifies any place action that is not final of a movable object on another object using an actuator. For example, \verb|Place(B, T, handL)| specifies placing the blue block on the table using the left hand. This move changes the position of the object (e.g., the red block) to be on top of the support object (e.g., an empty location on top of the table) while clearing its previous location. The rotation again is not handled at the symbolic planning stage.

\verb|Fix(Obj, Supp_Obj, Act)| specifies any place action that is final of a movable object on another object using an actuator. For example, \verb|Fix(G, R, handR)| specifies final fixation of the green block on top of the red block using the right hand. This move changes the position of the object (e.g., the green block) to be on top of the support object (e.g., red block) while clearing its previous location. Fix action is always final -- the object isn't moved after.

Given a pair of stack configurations as input, we wish to find sequences of moves (symbolic plans) that transform the initial stack to the target stack. We used Monte Carlo tree search (MCTS) to find satisfying sequences by branching the search tree using the three move types, the three objects, the four support objects, and the two actuators. Our pruning algorithm was efficient to a certain extent -- for example, if an object is already grasped, we did not branch the grasp move on it again. We also imposed a condition to produce a specialized set of solutions which we labeled as the efficient set, leaving the label inefficient for the universal set of solutions. To produce the efficient set, we would only branch the search tree to a \verb|Place(Obj,.,.)| if the \verb|Fix(Obj,.,.)| was not currently available for the block. We increased the maximum length of move sequences until no new unique solutions could be found. 

After a sequence was deemed satisfactory, we assigned integral timestamps to each of the abstract moves that it is composed of. These timestamps indicated the discrete-time values that an abstract move should be executed at. The assignment was done in a way to allow the execution of as many concurrent moves as possible. Of course, when a solution is one-handed, only one move can be executed at a time, thereby each abstract move must be assigned a separate timestamp. However, with two-handed solutions, different blocks can be concurrently actuated by different hands. Example symbolic plans for a pair of initial and target stack configurations are shown in Fig~\ref{fig:model}.

We assigned a complexity score to every symbolic solution generated, denoted $s_{i,j}$ where $i$ indexes problems and $j$ indexes its solutions. The score for a sequence is equal to the discrete-time that this sequence takes to terminate. 

\subsubsection{Geometric solver}
The geometric solver can be thought of as compiling a symbolic plan to manipulation trajectories of actuators and movable objects. It is based on a hierarchical optimization procedure for combined task and motion planning where the tasks come from the symbolic plan. Conditioned on the symbolic plan, the geometric solver generates a number of equality and inequality constraints that need to be met by the optimization procedure. These constraints are solved using an optimization package~\cite<k-order motion optimization framework, KOMO>{toussaint2014newton} that can handle long-distance dependencies such as the dependencies between actuator and object trajectories across time steps. Due to space limits, we cannot provide any further the details of KOMO and the logic-geometric programming framework \cite<but see>{toussaint2015logic,toussaint2014newton}. Snapshots of example manipulation trajectories generated by this optimization procedure for a pair of initial and target stack configurations are shown in Fig~\ref{fig:model}.



\subsection{Physical stability inference}
Because the geometric solver only considers kinematics and not the physical dynamics of the scene, it can find solutions that have physically unstable intermediate steps. Inspired by~\cite{battaglia2013simulation}, we infer whether a given intermediate configuration is stable by physically instantiating it in a physics engine (PhysX) and measuring the total kinetic energy over a total simulation duration of 1 sec with a burn-in period of 100 msecs. We reject a solution if the total kinetic energy exceeds an empirically determined threshold of 0.1 joules.

Similar to the complexity score for the symbolic solutions, we assigned an approximately metabolic cost score to every full model solution found (that is, solutions after the physical stability inference step), denoted $f_{i,j}$ where $i$ indexes problems and $j$ indexes its solutions. This score captures the extent to which a particular plan requires effort to execute. The score starts with the symbolic complexity score, $s_{i,j}$, but adds two more quantities: \textit{(1)} an extra cost of 0.5 for moves involving multiple blocks (e.g.,  actuating--i.e., grasping, placing or fixing-- the red block while the blue block rests on top of it), and \textit{(2)} an extra cost of 0.5 for moves that result in an intermediate physically unstable configuration from which the solver can recover to reach the correct stable configuration (e.g., moving the yellow block while the red block is leaning on it, and subsequently moving the red block). 




\subsection{Simulations and results}

\begin{figure}[t]
\centering
\includegraphics[width=.8\linewidth]{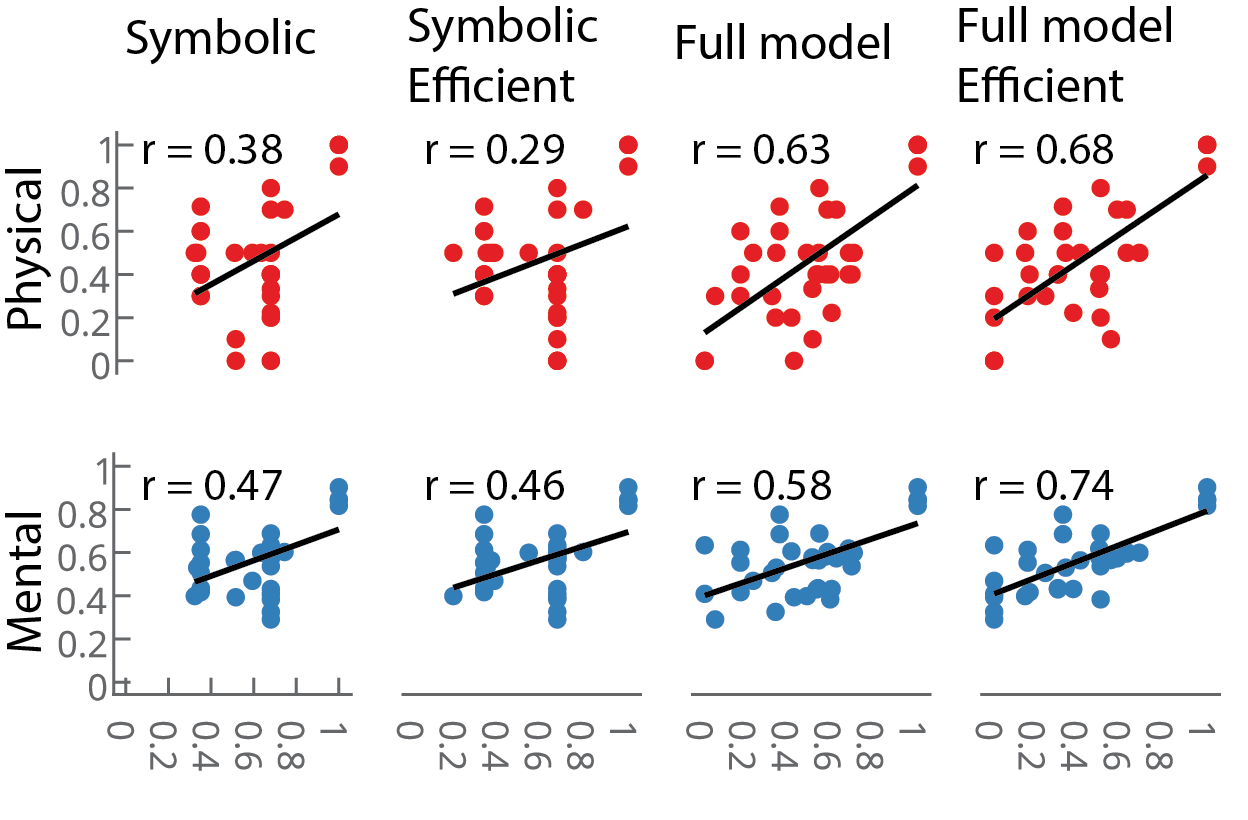}
\vspace{-0.4cm}
\caption{Scatter plots showing the relationship between different versions of the model (columns) and participants' actions in the lab (top), or mental simulations online (bottom). \emph{Note}: 1 = definitely one hand, 0 = definitely two hands.
}
\label{fig:model_results}
\vspace{-0.6cm}
\end{figure}

In addition to our full model, we also considered a lesioned model which leaves out the physical inference component.
We assume that people aim to reach their goal efficiently. Hence, we assume that sequences with higher complexity scores or metabolic costs are less likely to be chosen (in the lab) or simulated (online) than those with lower complexity scores or costs. For a given problem $i$, we obtain the probability of choosing one-hand based on the symbolic complexity scores in the following way
$
\frac{\sum_{j\in one-hand\ solutions}e^{-s_{i,j}}}{\sum_{j\in all\ solutions}e^{-s_{i,j}}}
.$ This means that the model is more likely to choose a one-hand solution the lower the cost of one-hand solutions are relative to all possible solutions.For the full model, the probability of choosing one-hand, Pr(One-hand), is calculated identically but using the full model scores, $f_{i,j}$. 

Overall, we found that the model accounted well for the data (see Fig.~\ref{fig:model_results}). In particular, we found that both physical stability inferences and efficiency were necessary to account for participants' judgments in Experiment 2 ($r=.74$, comparisons to symbolic-efficient, symbolic-inefficient and full-model-inefficient $p<.05$ using direct hypothesis testing with the bootstrap samples).

Similarly, in Experiment 1, we found that physical stability inferences were necessary to best explain participants' behavior (with $r=.68$ of the full model compared to $r=0.63$ of a model that doesn't take into account efficiency). But we did not find a statistical difference between using only the efficient solutions versus all solutions ($p=.06$).


\section{General Discussion}
We presented a novel paradigm -- the stack re-configuration problems -- and studied people's solving these problems in the laboratory (Experiment~1) and mentally simulating what they think a person would do (Experiment~2).
We found that participants' judgments about whether they think a person used one or two hands to get from the initial to the target configuration correlated well with participants' actual behavior in the lab. 

In order to explain participants' behavior, we developed a computational model that flexibly combines a symbolic, geometric, and physical representation of the scene. It efficiently plans over this representation by first forming a symbolic plan, trying to execute the plan using a geometric solver, and then checking whether the plan was feasible by consulting a physics simulation engine to make sure that each move resulted in a physically stable configuration. 

The full model accounts well for participants' actions as well as mental simulations. A model that does not take into account the efficiency of different plans fares worse (particularly when trying to explain mental simulations). Moreover, it is crucial to consider how much effort different plans would take into account well for participants' actions and judgments. Participants chose to use two hands only when a one-hand solution would have required considerably more effort. 



A striking aspect of problem solving is that it demands flexible systems that can operate with very little training opportunity, leading many researchers to emphasize the role of common-sense reasoning and model-building as the building blocks of human problem solving~\cite{johnson2005mental}. We find such flexibility and data efficiency in stark contrast with some of the main approaches to artificial intelligence today, in particular to deep learning~\cite{silver2016predictron}. These approaches require huge amounts of data, yet their generalization capacity is limited in contrast to human's flexibility. Turning these data-hungry approaches to flexible problem solvers is a substantial challenge. This paper makes a few (block) moves in this direction.

\fontsize{9}{11}\selectfont{
\noindent \textbf{Acknowledgments} This work was supported by the Center for Brains, Minds \& Machines (CBMM), funded by NSF STC award CCF-1231216 and by an ONR grant N00014-13-1-0333.
}

\renewcommand\bibliographytypesize{\small}
\bibliography{TobisPapers}
\bibliographystyle{bpacite}

\def\thebibliography#1{\section*{References}
\fontsize{6}{6}\selectfont
 \list
 {[\arabic{enumi}]}{\leftmargin \parindent
	 \itemindent -\parindent
	 \itemsep 0ex plus 1pt
	 \parsep 0.2ex plus 1pt minus 1pt
	 \usecounter{enumi}}
	 \def\newbrick{\hskip .11em plus .33em minus .07em}
	 \sloppy\clubpenalty4000\widowpenalty4000
	 \sfcode`\.=1000\relax}

\end{document}